\documentclass[letterpaper]{article}
\usepackage{iccc}
\usepackage{caption}
\usepackage{graphicx}
\usepackage{times}
\usepackage{helvet}
\usepackage{courier}
\usepackage{url}
\title{Automatically Detecting Amusing Games in Wordle}
\author{Ronaldo Luo, Gary Liang, Cindy Liu, Adam Kabbara,\\ {\Large \bf Minahil Bakhtawar, Kina Kim, Michael Guerzhoy}\\
Division of Engineering Science\\
University of Toronto\\
Toronto, Ontario, Canada\\
\{ronaldo.luo, esgary.liang, cindyjy.liu, adam.kabbara, minahil.bakhtawar\}@mail.utoronto.ca,\\ kina19131@gmail.com, guerzhoy@cs.toronto.edu\\
}

\begin{document} 
\maketitle
\begin{abstract}
\begin{quote}
    We explore automatically predicting which Wordle games Reddit users find amusing. 

    We scrape approximately 80k reactions by Reddit users to Wordle games from Reddit, classify the reactions as expressing amusement or not using OpenAI's GPT-3.5 using few-shot prompting, and verify that GPT-3.5's labels roughly correspond to human labels.
    
    We then extract features from Wordle games that can predict user amusement. We demonstrate that the features indeed provide a (weak) signal that predicts user amusement as predicted by GPT-3.5.
    
    Our results indicate that user amusement at Wordle games can be predicted computationally to some extent. We explore which features of the game contribute to user amusement. 

    We find that user amusement is predictable, indicating a measurable aspect of creativity infused into Wordle games through humor.
\end{quote}
\end{abstract}

\section{Introduction}

In this paper, we explore computationally predicting user amusement at games of Wordle. 

We scrape user reactions to games of Wordle from the \texttt{r/Wordle} subreddit, and use few-shot GPT-3.5 classification in order to rate the reactions to specific games as expressing amusement or not. We verify that GPT-3.5 roughly corresponds to human perceptions of amusement expressions.

We extract features from the game that we predicted would correspond to amusement: we count inherently funny words, as well as indicators of skill/luck. 
We show that our features weakly predict amusement reaction by users as predicted by GPT-3.5.

To the extent that the humor arises from users engaging with the game creatively, we report a measurement of the amount of creativity (rather than pure problem-solving) in the game (although of course humor can be a legitimate target for problem-solving).

\subsection{Prior work}
\cite{zaidi2024predicting} show that perception of move brilliance in chess can be predicted using features of the game trees. We focus on amusement rather than perception of brilliance, although those might be related. Rather than rely on annotations in chess (\texttt{"!!"} for ``brilliant move"), we must rely on GPT-3.5 assessment of text comments. This work, like us, measures human perception of game moves.

\section{Dataset}

We scrape a dataset of 80,000 combinations of Wordle games posted to \texttt{r/Wordle} and replies to those Wordle games. The replies often, though not always, are reactions to the Wordle game in the original post.

Five authors manually annotated 15 reactions from the scraped data whether they expressed amusement or not on a scale of 1-5. We then reviewed our separate annotation to converge on a common standard. After, the five authors separately annotated 85 more reactions. We achieved a weak to moderate agreement (see Table~\ref{tab:kappa_matrix}) on the variable $I(rating > 2)$.

\begin{table*}[h!]
\centering
\begin{tabular}{lcccccc}
\hline
 & Rachel & Mason & Clara & George & Alex & GPT \\
\hline
Rachel & 1.000000 & 0.312346 & 0.431938 & 0.372142 & 0.462562 & 0.398427 \\
Mason & 0.312346 & 1.000000 & 0.657948 & 0.828974 & 0.489270 & 0.158416 \\
Clara & 0.431938 & 0.657948 & 1.000000 & 0.572435 & 0.562232 & 0.242574 \\
George & 0.372142 & 0.828974 & 0.572435 & 1.000000 & 0.562232 & 0.158416 \\
Alex & 0.462562 & 0.489270 & 0.562232 & 0.562232 & 1.000000 & 0.268354 \\
GPT & 0.398427 & 0.158416 & 0.242574 & 0.158416 & 0.268354 & 1.000000 \\
\hline
\end{tabular}
\caption{Cohen's Kappa for (pseudonymous) authors and few-shot GPT. GPT prompt asked ``amusing or not", authors 1-5 ratings thresholded at 2. The table shows weak to moderate agreement\label{tab:kappa_matrix}}

\end{table*}

GPT-3.5-turbo was used using few-shot classification to classify 80,000 comments. GPT-3.5-turbo achieved weak-to-moderate agreement with the authors (see Table~\ref{tab:kappa_matrix}). The prompt is displayed in the Appendix.

\section{Method}
\subsection{Feature: Luck or Skill}
For each game, we compute indicators of luck and skill, which we hypothesize are related to amusement, as well as the inherent funniness of words.

The indicators of skill and luck are based on computing the reduction in the number of guesses that are still logically possible after a play and the length of the game. At each game state, one can compute the number of guesses that are consistent with the current state. We compute average/maximum reduction in the number of possible guesses after a play, the magnitude of the last reduction (to 0) in the number of possible guesses after a play, and the maximum/average/last Levenshtein distance and distance in GloVe~\cite{pennington2014glove} space between guesses. The motivation for the latter two is that it may be harder for humans to guess a word that is further away from the previous guesses.

\subsection{Feature: Intrinsically Funny Words}

A linear regression model with L2 regularization is trained to predict the ``intrinsic funniness" of words. The data used to train the model are 4997 words (4858 words after feature extraction) with humor rating ranging from 27.31 to 100~\cite{engelthaler2018humor}.

To train the model, each word in the dataset is represented by 19 different features: cosine distance to category-defining vector (CDV) of six categories (i.e. sex, party, insult, profanity, body function and animals), the Boolean value of whether the word contains the letter \verb;k;, the log frequency of word, the log average letter probability, the log average phoneme probability of the word, ratio of log average letter probability and log average phoneme probability, the Boolean value of whether the word contains the phoneme \verb;/u/;, the cosine distance to CDV of eight words ending with consonant+\verb;"le";, the valence (how appealing or undesirable the word makes people feel) $\times$ arousal (amount of emotional activation of word), valence $\times$ arousal $\times$ dominance (how controlling the word's referent is), arousal $\times$ dominance, valence, arousal, and concreteness (how tangible the word is). Those features are shown in \cite{westbury2019wriggly} to have  significant linear relationship with funniness.

First, the category defining vectors (CDVs) of six categories (i.e. sex, party, insult, profanity, body function and animals) that are related to funniness are computed \cite{westbury2019wriggly}. Those vectors are calculated by first getting 19 words most related to those categories as described in the Appendix A of \cite{westbury2019wriggly} and finding their average word2vec embedding; then we find cosine similarity between words within the dataset and the computed average embedding to find the average embedding of the first 100 words with greatest similarity \cite{westbury2019wriggly}. 

The CDV of the Cons+\verb;"le"; words are the average of word2vec embeddings of the words gaggle, jiggle, tinkle, waddle, wiggle, wriggle, gobble and nibble. The reasoning for using that is described in \cite{westbury2019wriggly}: Cons+"le" words with humor rating above 2 SD that do not carry an evident semantic relation to humor.

The frequency of a word is determined using data from~\cite{Tatman_2017}. The letter frequency of a word determined using data from~\cite{Practical_Cryptography}.

The pronunciation of a word is gathered from Carnegie Mellon University (CMU) Pronouncing Dictionary \cite{CMU_in_IPA}
and the frequency of each phoneme is used from data in \cite{CMLOEGCMLUIN_2012}.

The valence, arousal, dominance, and concreteness of a word are gathered from computational estimated of \cite{hollis2017extrapolating}.

After training a linear regression model using cross-validation, the root mean squared test error in predicting the funniness of words is $7.6717$ (over a range of $72.3$ of possible scores) and $R^2$ of 0.37022, implying we have a reasonable approximation of word funniness. The histogram and scatter plot for the distribution of original and predicted humor rating is displayed below in Fig.~\ref{fig:scatter} and Fig.~\ref{fig:histogram}.

\begin{figure}[h!]
    \centering
    \includegraphics[width=0.4\textwidth]{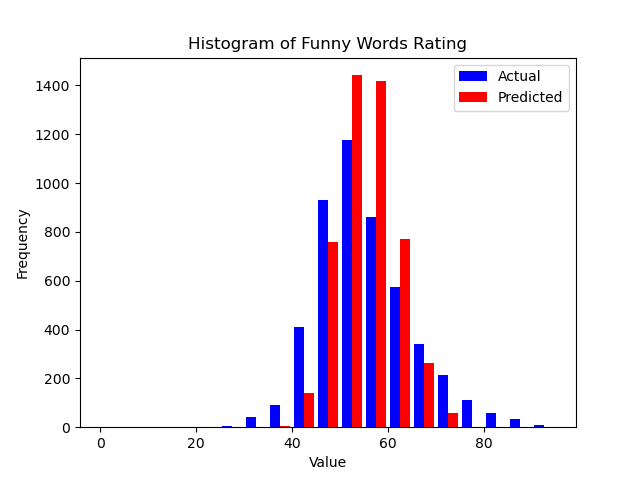}
    \caption{Actual and predicted humor ratings of the 4858 words in the training set}
    \label{fig:histogram}
\end{figure}
\begin{figure}[h!]
    \centering
    \includegraphics[width=0.4\textwidth]{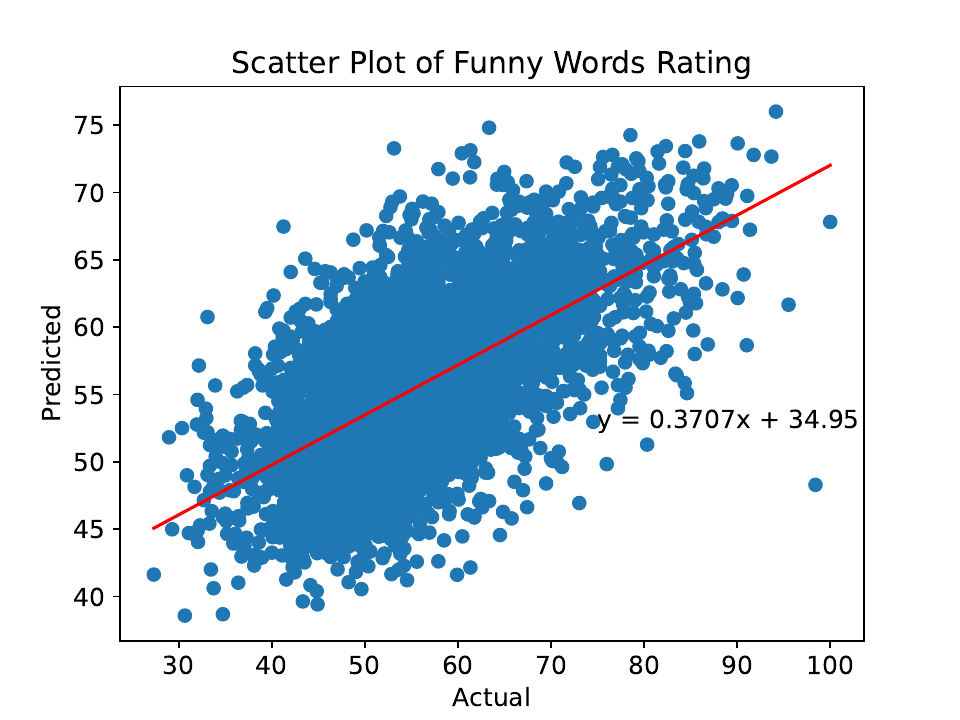} 
    \caption{Actual and predicted humor ratings of the 4858 words in the training set. }
    \label{fig:scatter}
\end{figure}

\subsection{Measure Funniness from Features}
Features showing skill or luck in the game may contribute to amusement reaction.

We hypothesize that semantic distance (as measured in GloVe space) and Levenshtein distance between guesses indicates skill or luck; it may be harder to pick a word that is further away in orthographic and semantic space~\cite{liang2024semantic}, and such a word may look more surprising.

\subsubsection{Levenshtein Distance}
Levenshtein distance measures the minimum number of single-character edits (insertions, deletions, or substitutions) required to transform one word into another~\cite{levenshtein1966binary}. In the context of gameplay, this metric quantifies the structural similarity between successive player guesses. A smaller Levenshtein distance indicates that a player is making guesses that are more similar to their previous attempts, potentially reflecting a bias toward familiar letter combinations or a cognitive tendency to minimize effort by sticking close to prior guesses. A larger Levenshtein distance between moves might imply the user is having fun with different guesses, or is possibly trying to play optimally.

\begin{table*}[ht!]
\centering
\captionsetup{width=0.8\linewidth}
\begin{tabular}{lrrrr}
\textbf{Coefficients} & \textbf{Estimate} & \textbf{Std. Error} & \textbf{z value} & \textbf{Pr($>|z|$)} \\ 

(Intercept)                    & 0.001321  & 0.010256 &  0.129 & 0.8975      \\ 
num\_possible\_guesses\_reduction\_max    & 0.017831  & 0.034160 &  0.522 & 0.6017      \\ 
num\_possible\_guesses\_reduction\_mean   & 0.073592  & 0.035754 &  2.058 & 0.0396 *    \\ 
num\_possible\_guesses\_reduction\_last   & 0.067377  & 0.015854 &  4.250 & 2.14e-05 *** \\ 
levenshtein\_distance\_max                & -0.041227 & 0.021723 & -1.898 & 0.0577 .    \\ 
levenshtein\_distance\_mean               & -0.227407 & 0.033649 & -6.758 & 1.40e-11 *** \\ 
levenshtein\_distance\_last               & 0.153290  & 0.024937 &  6.147 & 7.89e-10 *** \\ 
glove\_distance\_max                      & -0.028112 & 0.032958 & -0.853 & 0.3937      \\ 
glove\_distance\_mean                     & 0.021125  & 0.037529 &  0.563 & 0.5735      \\ 
glove\_distance\_last                     & 0.006977  & 0.021547 &  0.324 & 0.7461      \\ 
intrinsic\_humor\_of\_words\_max          & -0.012670 & 0.019760 & -0.641 & 0.5214      \\ 
intrinsic\_humor\_of\_words\_mean         & 0.132882  & 0.020178 &  6.586 & 4.53e-11 *** \\ 
intrinsic\_humor\_of\_words\_last         & 0.012456  & 0.013766 &  0.905 & 0.3656      \\ 
num\_possible\_guesses\_length            & -0.081160 & 0.016345 & -4.965 & 6.85e-07 *** \\ 

\end{tabular}
\caption{Logistic regression coefficients for predicting amusingness from features, their estimates, standard errors, z-values, and p-values. Significance codes: 0 ‘***’, 0.001 ‘**’, 0.01 ‘*’, 0.05 ‘.’, 0.1 ‘ ’ 1.}
\label{tab:coefficients}
\end{table*}

\subsubsection{Semantic Distance (GloVe)}
The semantic distance between guesses is computed using GloVe embeddings~\cite{pennington2014glove}. GloVe distance is calculated as the negative cosine similarity between these vectors, allowing us to quantify how conceptually similar two guesses are.

A smaller GloVe distance suggests that a player's guesses are semantically related, possibly indicating that the user is not being creative with moves, or possibly that the user is enjoying making semantically-related plays.

\subsubsection{Drop in number of possible guesses}
A drop in the number of possible guesses left may indicate either skill or luck, contributing to amusement.

The following features are computed:

\begin{itemize}
    \item \{Maximum/average/last\} $\times$ \{drop in number of guesses, GloVe space distance between guesses, funniness of word for guesses\} (9 features)
    \item Drop in number of guesses, GloVe space distance to previous guess, funniness of word for guesses ($6\times 3$ features, padded with 0s as necessary)
    \item Number of moves in the game
\end{itemize}

\subsection{Model and Performance}
We trained logistic regression, regularized logistic regression, and fully-connected feedforward networks with ReLU activations with the architectures 
\begin{verbatim}
    NFEAT-10-10-1
    NFEAT-10-1
    NFEAT-100-10-1
\end{verbatim}

We used a 60-20-20 train/validation/test split, and explored regularization. We subsampled our dataset so that the labels are distributed as 50/50 amused/not-amused (our initial distribution was 75\% not amused/25\% amused)

We found that neural networks performed worse than logistic regression. We found that adding in features per individual word rather than aggregate features did not change performance on the validation set, but did lead to overfitting, especially with the neural network architectures.

Performance on the test set, for all the settings described above, was $54\% \pm 0.5\%$, using all the settings above. 

We therefore report results using logistic regression.

\section{Results}
We report results using logistic regression, with only aggregate features used to predict amusement reaction.

We achieve correct classification rate of $54.5\%$ with a baserate of $50\%$, on a dataset of size 15,000. The result is therefore modest but significant: it is possible to predict amusement reaction from the game.

We normalize the features in order to demonstrate which features are more important for predicting amusement reactions. The results are in Table~\ref{tab:coefficients}. We observe that GloVe distance does not seem to affect prediction of perception of humor. As predicted, the intrinsic humor of words does affect perception of amusement.

Unsurprisingly, shorter games predict perception of amusement, perhaps because of a show of skill or luck.

Large reductions in the number of possible guesses also correlate with a show of skill or luck and predict perception of amusement.

The effect of the Levenshtein distance between the guesses is interesting: the last two guesses being far apart is predicted to be more amusing, but the aggregate measures of Levenshtein distance seem to lead to less predicted amusement the larger they are. That is perhaps because a last game where the guess is orthographically very different from the previous guess shows skill or luck in a more obvious way.

Note that the significant p-values are generally very small, and would be robust to a Bonferroni correction.

For reference, an increase of one standard deviation from the mean in the Levenshtein distance between the two last guesses increases the probability of an amused reaction by about 4\%, and an increase of one standard deviation from the mean in the reduction in the number of possible guesses between the last and the second last guess increases the probability of an amused reaction by about 2\%. Those effects are small but statistically significant.

We observe that a univariate logistic regression where the only input is the number of guesses (\verb;num_possible_guesses_length;) results in a correct classification rate that is smaller by only $0.2\%$ from the full model. While the models are different (Chi-squared test p-value $< 0.001$), the difference is small.

Interestingly, when we run a linear regression, our features predict the number of guesses with an $R^2$ of $60\%$, partially explaining our observation -- the number of guesses is correlated with the other features.

\section{Conclusions}
We show that perception of amusement at Wordle games can be predicted, to a very modest extent, from the games.

We note that our rating of comments as expressing amusement is likely very imperfect: while the Cohen's Kappa coefficient is always positive, both the interrater agreement and agreement between the different authors and between the authors and GPT-3.5 is quite modest.

We hypothesize that better agreement on what it means for a comment to express amusement would result in better predictive performance.



\bibliographystyle{iccc}
\bibliography{iccc}

\appendix

\section{Few-shot prompt for GPT-3.5-turbo \label{sec:prompt}}
\tiny
\begin{verbatim}
    def classify_funny(comment):
    response = openai.ChatCompletion.create(
        model="gpt-3.5-turbo",
        messages=[
            {
                "role": "system",
                "content": (
                    "You are a helpful assistant that classifies humor in 
                    comments. Humor is defined as the extent that the commenter
                    is amused by the Wordle game."
                    "Rate each comment using a binary convention: '1' if the
                    comment is funny and '0' if the comment is unfunny. "
                    "Your response should ONLY be '0' or '1'. Do not include
                    any additional explanation."
                ),
            },
            {
                "role": "user",
                "content": (
                    "I will give you examples of comments and their binary
                    humor labels. Use these examples to classify new comments 
                    using the same convention:\n"
                    "1 -> Funny\n"
                    "0 -> Unfunny\n\n"
                    "Examples:\n"
                    f"{examples_text}\n\n"
                    "Now, classify the following comment strictly as
                    '0' or '1':\n"
                    f"Comment: '{comment}'"
                ),
            },
        ],
        max_tokens=2,
        temperature=0
    )
\end{verbatim}

\end{document}